\documentclass[letterpaper,10pt,conference]{ieeeconf}
\IEEEoverridecommandlockouts
\usepackage{amsmath}
\usepackage{amssymb}
\usepackage{graphicx}
\usepackage{booktabs}
\usepackage{array}
\usepackage{xcolor}
\usepackage{cite}
\usepackage{url}
\usepackage{balance}
\usepackage{tikz}
\usetikzlibrary{arrows.meta}
\graphicspath{{figures/}}

\newcommand{\result}[1]{#1}
\usepackage[normalem]{ulem}
\newif\ifshowchanges \showchangesfalse
\newcommand{\cadd}[1]{\ifshowchanges{\color{teal}#1}\else#1\fi}
\newcommand{\cdel}[1]{\ifshowchanges{\color{red}\sout{#1}}\fi}
\newcommand{\cdelm}[1]{\ifshowchanges{\color{red}#1}\fi}
\newif\ifshowfinal \showfinalfalse
\newcommand{\fadd}[1]{\ifshowfinal{\color{magenta}#1}\else#1\fi}
\newcommand{\fdel}[1]{\ifshowfinal{\color{red}\sout{#1}}\fi}
\newcommand{\fdelm}[1]{\ifshowfinal{\color{red}#1}\fi}
\newif\ifshowrefs \showrefsfalse
\newcommand{\radd}[1]{\ifshowrefs{\color{blue}#1}\else#1\fi}
\newcommand{\rdel}[1]{\ifshowrefs{\color{red}\sout{#1}}\fi}
\newcommand{\rdelm}[1]{\ifshowrefs{\color{red}#1}\fi}
\newif\ifshowprof \showproffalse
\newcommand{\padd}[1]{\ifshowprof{\color{orange}#1}\else#1\fi}
\newcommand{\pdel}[1]{\ifshowprof{\color{red}\sout{#1}}\fi}

\title{\LARGE \bf
Dual-Rate Force--Image Control with Model-Based Orientation Limits for Robotic Ultrasound
}
\author{Tyler Foster$^{1}$, Qiang Zhang$^{1,2}$, \IEEEmembership{Member, IEEE}, A B M Tahidul Haque$^{1}$, and Anh Thu Nguyen$^{3}$%
\thanks{$^{1}$Tyler Foster, Qiang Zhang, and A B M Tahidul Haque are with the Department of Mechanical Engineering, The University of Alabama, Tuscaloosa, AL 35487, USA.
$^{2}$Qiang Zhang is also with the Department of Chemical \& Biological Engineering, The University of Alabama, Tuscaloosa, AL 35487, USA. $^{3}$Anh Thu Nguyen is with the Department of Computer Science, The University of Alabama, Tuscaloosa, AL 35487, USA.
Corresponding author: Qiang Zhang, {\tt\small qiang.zhang@ua.edu}.}%
}
\begin{document}
\maketitle
\thispagestyle{empty}
\pagestyle{empty}
\begin{abstract}
Robotic ultrasound couples a high-rate contact-force loop with slower, delayed image feedback, so image-guided ultrasound probe rotation can perturb contact force before the resulting image response is observed. We derive a closed-form orientation-rate limit that bounds the modeled rotation-induced estimated-force excursion over a finite horizon while accounting for disturbance rejection by the fast force loop. The limit depends on local contact stiffness, force-loop gains, a conservative rotation-to-force gain bound, the excursion budget, and the prediction horizon. We implement this model in a dual-rate controller with timestamp-based delay reconstruction and joint-torque-based force estimation, and evaluate it on a curved gelatin phantom using paired controller comparisons and component ablations. Relative to unconstrained image guidance, the proposed rate-limited controller reduced first-second root-mean-square (RMS) estimated-force error by 0.40 N while increasing cue-convergence time by 0.94 s. A fixed rate cap near the analytically predicted ceiling produced no resolvable difference in force error and converged 0.32 s faster, indicating that the principal practical value of the model is the rate-design rule rather than online prediction. Delay reconstruction had no resolvable effect at the tested latency. A single-subject popliteal scan demonstrated feasibility, although the image cue was noise-limited on heterogeneous tissue.

\end{abstract}
\section{Introduction}
\label{sec:introduction}

Brightness mode (B-mode) ultrasound provides real-time, radiation-free imaging at comparatively low cost, but its image quality depends strongly on how the probe interacts with tissue. A sonographer continuously coordinates probe position, orientation, and contact force while responding to surface geometry, tissue compliance, and the image itself\cdel{, so}\cadd{;} robotic ultrasound is \cadd{therefore }a coupled sensing-and-control problem rather than a trajectory-following task. Reviews accordingly identify scan-path generation, probe--tissue force regulation, and image-quality or orientation optimization as the central components of autonomous systems \cite{li2021overview,salcudean2022,jiang2023review}, and recent surveys \radd{\rdelm{\cite{bi2024,han2025clinical,munir2025}}\cite{han2025clinical,munir2025} note open gaps, among them }\rdel{note }that orientation-related interaction forces are often simplified and that real-time image feedback is still rarely integrated into online scan optimization\rdelm{ \cite{bi2024,munir2025,han2025clinical}}\radd{~\cite{munir2025}}.

Image features extracted directly from ultrasound can close the robot-positioning loop \cite{abolmaesumi2002}, and visual servoing on ultrasound confidence has been used to improve \rdel{acoustic coupling and}\radd{image quality through} probe orientation \cite{chatelain2015}. 
Interaction-force information \radd{\cite{jiang2020normal,jiang2021force} and probe-mounted distance sensing \cite{ma2022asee} }\rdel{has}\radd{have} been used to estimate the local surface normal and regulate orientation\rdelm{ \cite{jiang2020normal,jiang2021force,ma2022asee}}, explicit force/moment control improves interaction under heterogeneous contact \cite{piedra2026}, ultrasound feedback has enabled autonomous screening of tubular structures \cite{jiang2022tubular}, and image-quality estimates have been used to adapt the commanded force \cite{akbari2021,zakeri2025}. Image and force sensing are therefore established as complementary\pdel{, and combining them is no longer, by itself, a contribution to control}. Learning-based methods have extended autonomy to navigation, compliant scanning, and complete clinical workflows \cite{hase2020,ning2021rl,jiang2025ultrabot}, including policies trained under explicit force constraints \cite{duan2025safe}; this work instead keeps a physically structured force loop and uses the image only as a measured cue.

A control issue that is less explicitly addressed is the multirate dynamic coupling between force regulation and image-guided probe correction. Robot state and wrench estimates update at high rates, whereas B-mode images are acquired, transferred, and processed at much lower rates and with non-negligible latency\cdel{; image-based force adjustment, for example, has been run with an image loop slower than the underlying robot loop \mbox{\cite{akbari2021}}}. The difficulty is more fundamental than asynchronous sensing alone\cadd{, because the two loops are coupled through the contact}. An orientation correction computed from an image changes probe indentation, the contact moment, and consequently the normal force before a new image becomes available. Conversely, a fast force regulator reacts to the normal displacement induced by the orientation update and can partially oppose the intended correction.\cdel{ Tissue stiffness, local curvature, and uncertainty in image sensitivity further modify this bidirectional interaction.} Dual-loop approaches have demonstrated image-dependent force adaptation and robust force tracking \cite{zakeri2025}, Li \emph{et al.}~\cite{li2026trajectory} incorporate trajectory-dependent feedforward and delay compensation for contact-force tracking, and Qian \emph{et al.}~\cite{qian2026symmetry} combine image guidance with biomechanics-informed force measurement. These approaches do not\pdel{, however,} treat the image-guided orientation loop and the high-rate force loop as a delayed multirate system whose cross-coupling is predicted and used in the controller.

This paper treats that coupling as the control problem. We propose a dual-rate orientation controller in which the slow image-guided loop accounts for both its own latency and its effect on the fast contact loop: the image cue is reconstructed to the present instant from the frame's capture timestamp; the force excursion that a candidate correction would produce is predicted from a conservative bound on the rotation-to-force gain; and the prediction credits the fast loop with the \cdel{part}\cadd{portion} of that excursion it will reject \cdel{before the next frame arrives}\cadd{within the prediction horizon}. \fdel{The last term}\fadd{This credit} separates the method from a one-step safety filter, which must assume the excursion persists and therefore admits a lower rotation rate. While the correction demand is saturated, the constraint reduces to a ceiling on the rotation rate that depends only on the contact stiffness, the force-loop gains, the gain bound, \fdel{and }the force budget\fadd{, and the prediction horizon}, so it can be evaluated before image-guided correction begins, once the local stiffness and a valid upper bound on rotation-to-force gain have been established.
The \pdel{controller runs on a }platform \pdel{with}\padd{has} \emph{no force/torque sensor}; contact force is estimated from joint torques. This raises the force noise floor and introduces model-dependent bias, so the force margins the controller must respect are small relative to the estimate's \pdel{own }error. The contributions are:

\vspace{-0.5mm}
\begin{enumerate}
    \item a closed-form design rule for the maximum image-guided orientation rate that limits the modeled rotation-induced force excursion over a finite horizon while accounting for force-loop rejection; 

    \item a paired hardware evaluation on a curved phantom against force-only, newest-frame, and fixed-rate-cap baselines, with ablations isolating the constraint, the fast-loop credit, delay reconstruction, and online sensitivity estimation, showing \fdel{that the model predicts the realized rotation rates, }that the ceiling sets the trade-off between cue-convergence time and force error, and that a fixed cap \fdel{at}\fadd{near} the predicted rate matches the online constraint\fadd{ in force error}\pdel{, together with}\padd{; and}

    \item \pdel{a model-free measurement of the rotation-to-force gain and a volunteer popliteal scan}\padd{a single-subject popliteal scan testing feasibility on living tissue}.
\end{enumerate}
\section{System and Sensorless Force Estimation}
\label{sec:system}

\subsection{Platform}
\begin{figure}[t]
\centering
\includegraphics[width=\columnwidth, trim=2.0in 0.85in 1.75in 0.85in, clip]{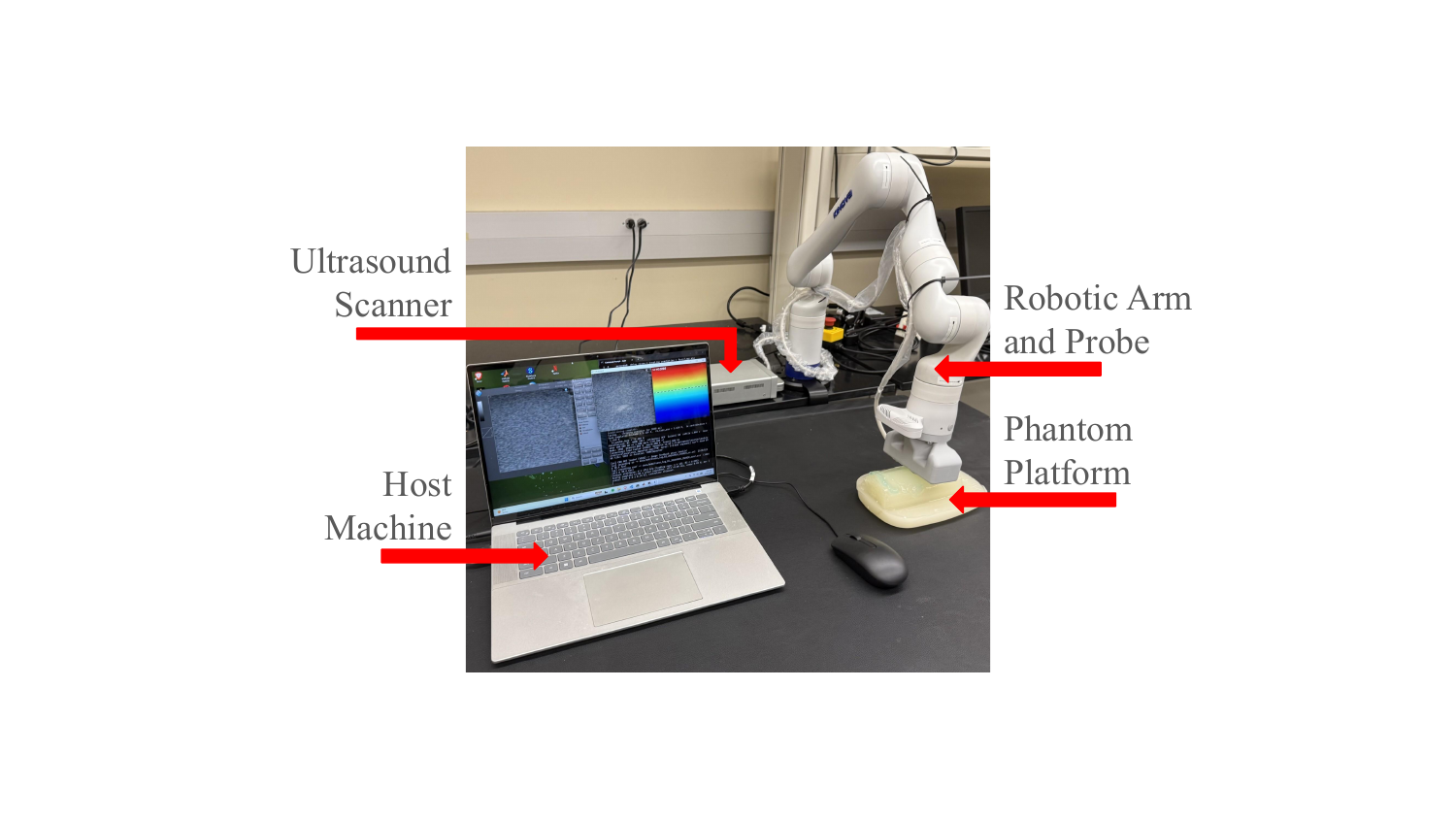}
\vspace{-8 mm}
\caption{Experimental setup\pdel{.}\padd{: a Kinova Gen3 six-degree-of-freedom arm holds a Telemed LF9-5N60-A3 linear probe in a printed holder against the gelatin phantom; the host machine runs the controller and shows the live B-mode stream and its confidence map.}}
\vspace{-3 mm}
\label{fig:setup}
\end{figure}

The platform (Fig.~\ref{fig:setup}) is a Kinova Gen3 six-degree-of-freedom manipulator carrying a Telemed \fdel{Artus}\fadd{LF9-5N60-A3} linear probe in a rigid printed holder. No force/torque sensor is present; the contact wrench is estimated from the torque sensors in each actuator. This removes the cost, cabling, and sterilization burden of a wrist-mounted \fdel{transducer}\fadd{sensor} at the expense of a higher noise floor and a model-dependent bias.

The arm is commanded in joint position at $f_f=1$~kHz through the manufacturer's low-level cyclic interface. A Cartesian velocity command $v_c\in\mathbb R^{6}$ is mapped to joint rates through a damped least-squares inverse $J^{\dagger}$ of the geometric Jacobian $J(q)\in\mathbb R^{6\times6}$ and integrated to the streamed position command, as
\begin{equation}
\dot q_c=J^{\dagger}(q)\,v_c,\qquad
q_c^{k+1}=q_c^{k}+\dot q_c\,\Delta t_f,
\label{eq:intcmd}
\end{equation}
with $q\in\mathbb R^{6}$ the measured joint coordinates and $\Delta t_f=1/f_f$. B-mode frames are captured from the scanner display at $f_I=30$~Hz, and each is tagged with its capture time. The scanner's internal acquisition-to-display delay was not resolved and is not compensated.

\subsection{Wrench Estimation from Joint Torques}
With $\tau_m\in\mathbb R^{6}$ the measured joint torques and $h_c\in\mathbb R^{6}$ the \fdel{contact wrench at the tool}\fadd{probe's wrench on the tissue}, the rigid-body model is
\begin{equation}
M(q)\ddot q+C(q,\dot q)\dot q+\tau_g(q)+\tau_r\cadd{+J^{\!\top}(q)\,h_c}=\tau_m\cdelm{+J^{\!\top}(q)\,h_c} ,
\label{eq:dyn}
\end{equation}
where $\tau_g$\padd{${}\in\mathbb R^{6}$} is the gravity torque and $\tau_r$\padd{${}\in\mathbb R^{6}$} \cadd{is }a joint-level residual. The gravity torque is computed by a quasi-static recursive Newton--Euler pass
\cite{featherstone2008} with link parameters from the manufacturer's model, the last
link's mass and center of mass updated for the probe and holder, and the gravity vector
corrected \cdel{for the base mounting tilt}\cadd{to its measured value at the base}. The \cdel{velocity- and acceleration-dependent}\cadd{acceleration- and velocity-dependent} (inertial
and Coriolis) terms are omitted. \fdel{Replaying}\fadd{Evaluated on} the logged trajectories \fdel{through}\fadd{with} the full model,
they contribute at most \result{1.2}~N to the regulated force in the first second after
rotation onset and \result{0.8}~N during the traverse for joint motion below 5~Hz
(\result{2.0} and \result{2.1}~N below 10~Hz), and do not change the sign or significance of any
\padd{controller }comparison\pdel{ in Table~\ref{tab:ablation}}\padd{ (Section~\ref{sec:results})}. Including them online would require
differentiating the \fdel{actuators' }velocity feedback, itself \cdel{a numerical derivative}\cadd{computed by the actuator\fadd{s}}; on the
logged data this inflates the computed terms \cdel{roughly tenfold}\cadd{at least fivefold} with noise.

The residual $\tau_r$ depends on joint position and on the direction of recent motion. It is modeled per joint by two branches, each a constant plus low-frequency harmonics of joint angle (periods of $45$--$290^\circ$) identified offline from free-space pose sweeps. The weight between branches evolves as a single stop \radd{(elastic--plastic) }operator of width $1^\circ$ \cite{brokate1996}, so the model blends linearly between branches over the first degree of travel after a reversal; it is a low-order model with no minor-loop memory. The estimated contact wrench is
\begin{equation}
\hat h_c=\bigl(J^{\!\top}(q)\bigr)^{\!+}
\bigl(\bar\tau_m-\tau_g(q)-\hat\tau_r\fdelm{(q)}\bigr)\cdelm{-h_0} ,
\label{eq:wrench}
\end{equation}
where $\bar\tau_m$ is $\tau_m$ after a per-joint second-order Butterworth filter with a $25$~Hz cutoff, $(\cdot)^{+}$ denotes the pseudoinverse, \fadd{and }$\hat\tau_r$ is the residual model\cdel{, and $h_0\in\mathbb R^{6}$ is a free-space tare acquired while the arm holds position immediately before contact. Because $h_0$ absorbs whatever residual the model leaves at that pose, it is preceded by a conditioning motion that walks each joint against and then along its press direction, so the tare is taken on the same branch that the subsequent press will traverse.}\cadd{.}

The regulated variable is the beam-axis component of the estimated contact force in a probe frame whose origin is the acoustic face,
\begin{equation}
F_n=\fdelm{-\,}e_3^{\!\top}\cadd{\bigl(}\,{}^{P}\!R_B\,\hat F_c\cadd{-f_0\bigr)} ,
\label{eq:fn}
\end{equation}
where $\hat F_c\in\mathbb R^{3}$ is the force \cdel{part}\cadd{component} of $\hat h_c$, ${}^{P}\!R_B\in SO(3)$ the base-to-probe rotation, \cadd{$f_0\in\mathbb R^{3}$ a free-space tare of ${}^{P}\!R_B\hat F_c$ acquired while the arm holds position immediately before contact, }and $e_3\in\mathbb R^{3}$ selects the beam axis, which points out of the face, so compression is positive.\cadd{ Because $f_0$ absorbs whatever residual the model leaves at that pose, it is preceded by a conditioning motion that walks each joint against and then along its press direction, so the tare is taken on the same branch that the subsequent press will traverse.}

\subsection{Achievable Accuracy}
Since \eqref{eq:wrench} has no independent force reference during operation, its accuracy bounds every downstream claim, and all force results are stated in estimate space. Static loading against a reference scale was repeated at eight arm poses, each holding five levels between $5$ and $20$~N under force control\cdel{ (the loop held every level to within $0.02$~N of its setpoint).}\cadd{; the loop held every level within $0.02$~N of its setpoint.} Over the $40$ contact dwells, the error of \eqref{eq:wrench} against the scale was $0.41$~N \padd{root-mean-square (}RMS\padd{)} with the residual model and $0.49$~N without it, \fdel{evaluated on the same dwells, }with maxima of $1.11$ and $1.27$~N; the model lowered the error at $30$ of the $40$ dwells. The mean error at each level lay between $-0.02$ and $-0.28$~N, but its size and sign varied with pose (from $-0.72$ to $+0.48$~N at $5$~N), so \pdel{it is not a gain error that }a single scale factor would \padd{not }remove\padd{ it}. After the joint reversal at retraction, the free-space reading was offset by $1.19$~N RMS with the model and $1.45$~N without it, the model over-correcting the reversal (mean $+1.08$~N against $-1.41$~N)\cadd{ but still reducing its magnitude}. These figures bound the estimate rather than calibrate it.
\section{Orientation Sensing from Ultrasound}
\label{sec:sensing}

\begin{figure}[t]
\centering
\includegraphics[width=\columnwidth]{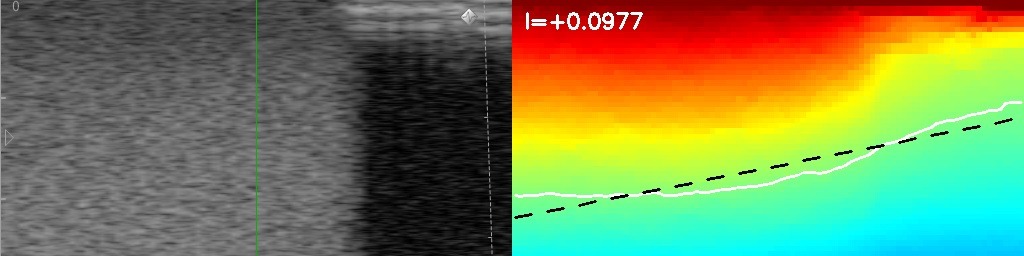}\\[-1pt]
{\footnotesize\makebox[0.5\columnwidth]{(a)}\makebox[0.5\columnwidth]{(b)}} 
\vspace{-6 mm}
\caption{\padd{Image cue on the phantom. (a)~B-mode frame; the right side of the image is darker. (b)~Its confidence map $c(r,\kappa)$ (blue $0$ to red $1$), with the depth $d(\kappa)$ at which confidence falls below $c_{\rm th}=0.5$ (white) and its least-squares line (dashed). The confidence threshold is reached at shallower depth on the right, so $d(\kappa)$ falls with $x_\kappa$ and, by \eqref{eq:cslope}, $I=+0.0977$. Upper half of the image shown.}}
\vspace{-3 mm}
\label{fig:cue}
\end{figure}

\cdel{When the probe deviates from the local surface normal, acoustic coupling degrades on one side of the aperture and image quality falls off from the transducer face downward on that side.}\cadd{Misalignment with the surface or partial loss of acoustic coupling lowers confidence on one side of the aperture, where image quality falls off from the transducer face downward.} We quantify this asymmetry with a \fadd{dimensionless }scalar cue $I$, \fdel{dimensionless, }with $I=0$ \fdel{corresponding to}\fadd{for} a laterally symmetric image.

The cue is computed from the ultrasound confidence map of Karamalis \emph{et al.}~\cite{karamalis2012}, which assigns to each pixel the probability that a random walker released there reaches the transducer before the far field, with edge weights derived from intensity gradients under an attenuation model. Writing $c(r,\kappa)\in[0,1]$ for the confidence at row $r$ and column $\kappa$, the map is obtained by solving a sparse Dirichlet problem on the pixel graph, here at $96\times96$ on a fixed crop of the displayed image\cdel{ (which includes the scanner's on-screen markings)} by direct factorization with the symbolic analysis performed once. The per-frame cost on the experimental hardware, including crop, resize, solve, and
feature extraction, was \result{22.4 $\pm$ 4.8}~ms (95th percentile \result{30.7}~ms;
\result{2}\% of frames exceeded the $33$~ms frame interval). Because the map is solved in
parallel with capture, the control signal always comes from the previous frame, or from
two frames back when a solve overruns. Each result is tagged with the capture time of the
frame from which it was computed\padd{, reaches the controller a median of $33.8$~ms after that capture, and is held for up to one frame interval}.

The cue uses the depth at which confidence falls below a threshold $c_{\rm th}$,
\begin{equation}
d(\kappa)=\frac{1}{N_r-1}\min\{\,r:\;c(r,\kappa)<c_{\rm th}\,\},
\label{eq:cdepth}
\end{equation}
with $N_r$ the number of rows, and takes its least-squares slope against normalized lateral position $x_\kappa\in[-1,1]$, negated so that a more confident left side gives a positive value\padd{ (Fig.~\ref{fig:cue})}:
\begin{equation}
I=-\,\Bigl(\textstyle\sum_\kappa x_\kappa d(\kappa)\Bigr)
\Bigl/\Bigl(\textstyle\sum_\kappa x_\kappa^{2}\Bigr).
\label{eq:cslope}
\end{equation}
\cdel{Because \eqref{eq:cslope} depends on the depth at which the reliable region ends rather than on the confidence level itself, a spatially uniform change in gain or in overall confidence leaves it unchanged.}\cadd{Because the map normalizes image intensity, a uniform change in scanner gain leaves it unchanged, and because \eqref{eq:cslope} is a slope, a change in depth common to all columns leaves $I$ unchanged.} Map parameters\radd{~\cite{karamalis2012}} were selected offline on recorded scans: \cadd{the depth attenuation }$\alpha=2.0$, \cadd{the edge-weight sharpness }$\beta=10.0$, \cadd{the lateral edge penalty }$\gamma=0.05$\cadd{\rdelm{~\cite{karamalis2012}}}, and $c_{\rm th}=0.5$.

The cue used for control, $I_f$, is $I$ passed through a second-order Butterworth low-pass filter with a $10$~Hz cutoff, stepped at $1$~kHz on the most recent frame value. Its group delay near DC, $\tau_f=\sqrt2/(2\pi\cdot10~\mathrm{Hz})=22.5$~ms, is compensated in the delay reconstruction of Section~\ref{sec:control}.

The cue presupposes that most of the aperture is coupled. If contact is lost across a large fraction of the face, $I$ remains well defined but no longer reports misalignment. This is a limitation of the cue rather than a condition the controller detects; the \fdel{excursion}\fadd{tilt} limit of Section~\ref{sec:control} bounds how far a misreported cue can drive the probe.
\section{Dual-Rate Control}
\label{sec:control}
\begin{figure*}[t]
  \centering
  \includegraphics[width=1.0\linewidth]{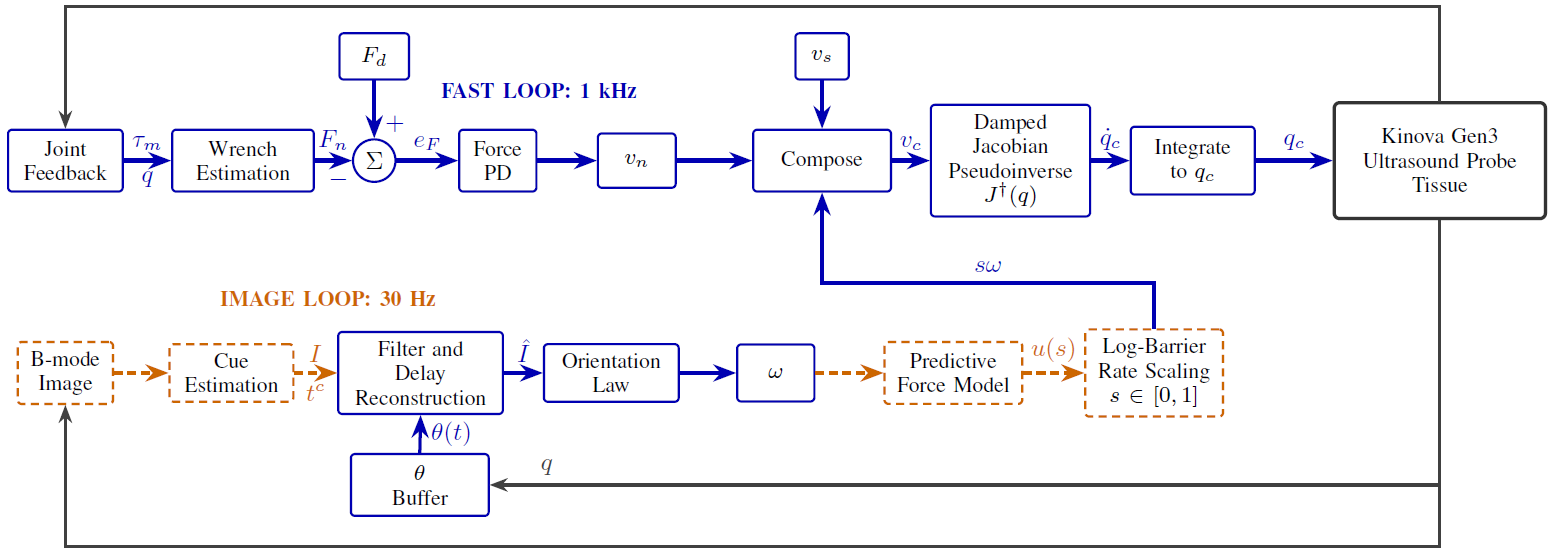}
  \vspace{-7 mm}
  \caption{Dual-rate control architecture. \fdelm{The $30$~Hz image loop reconstructs the cue to the present instant and scales the orientation command through the force-excursion constraint; the $1$~kHz force loop regulates the estimated contact force}\fadd{Dashed orange: once per image frame ($30$~Hz); solid blue: every $1$~kHz tick; gray: physical and sensor paths. The cue is filtered and reconstructed to the present instant, the orientation command is scaled through the force-excursion constraint, and the force loop regulates the estimated contact force}\cadd{; $v_s$ is the constant $5$~mm/s scan feed}.}
  \vspace{-3 mm}
  \label{fig:control_architecture}
\end{figure*}

\subsection{Fast Force Loop}
The force loop regulates $F_n$ to a constant reference $F_d$ by commanding beam-axis velocity at $1$~kHz\fadd{ (Fig.~\ref{fig:control_architecture})},
\begin{equation}
v_n=K_p e_F-K_d\dot F_n,
\qquad e_F=F_d-F_n,
\label{eq:pd}
\end{equation}
with $v_n$ saturated at $0.04$~m/s and the derivative taken on the measurement. No integral term is used: the plant integrates the velocity command, so the loop already rejects constant force disturbances. \cdel{Gains were tuned at the operating force (}\cadd{The gains (}$K_p=6.25\times10^{-3}$~m\,N$^{-1}$\,s$^{-1}$, $K_d=6.25\times10^{-5}$~m\,N$^{-1}$) \cadd{were }held fixed across every reported condition\cadd{, and a safety stop ends a run if $|F_n|$ exceeds $30$~N}.

Locally, the contact obeys $\dot F_n=K v_n+g_\theta\,\omega$, where $K$ is the contact stiffness, $\omega\padd{{}\in\mathbb R}$ the probe's rotation rate in the image plane, and $g_\theta$ the rotation-to-force gain defined below. With \eqref{eq:pd}, the force error then evolves as
\begin{equation}
\dot e_F=-a\,e_F-g'\omega,
\label{eq:arate}
\end{equation}
with $a=KK_p/(1+KK_d)$ the fast loop's disturbance-rejection rate and $g'=g_\theta/(1+KK_d)$. $K$ is identified once per contact from the loading ramp, by a least-squares fit of $F_n$ against face depth along the beam axis (from forward kinematics, \cdel{filtered like the torques}\cadd{low-pass filtered with the same Butterworth filter as the joint torques}) over the samples between $0.4F_d$ and touchdown\cadd{, when $F_n$ first comes within $10\%$ of $F_d$}; a fit with fewer than $100$ samples or $R^2<0.8$ is rejected in favor of a prior. The ramp precedes the scan feed, so surface elevation does not enter the fit. At the median per-contact stiffness of the comparison runs, $K=1133$~N/m (Section~\ref{sec:char}), the loop gives $a=6.6$~s$^{-1}$ and $1+KK_d=1.07$.

\subsection{Orientation Channel}
The orientation loop rotates the probe about the \fdel{in-plane}\fadd{elevational} axis of its face, changing its tilt $\theta\padd{{}\in\mathbb R}$ in the image plane (measured from the touchdown orientation), so as to drive the cue to zero,
\begin{equation}
\omega=-K_I\,\sigma\,I,\qquad
\sigma=\operatorname{sign}\!\left(S\right),
\label{eq:imglaw}
\end{equation}
with cue gain \cdel{$K_I>0$}\cadd{$K_I=40$~rad/s}\fadd{, $\sigma$ the sign of the cue sensitivity $S$ defined below,} and $\omega$ saturated at $\omega_{\max}$.\cdel{ The command is mapped to the Jacobian's reference point so that the face pivots in place rather than translating.} Net \fdel{angular excursion}\fadd{tilt} from the touchdown orientation is limited to $0.55$~rad\fadd{ ($31.5^\circ$)}\cdel{; beyond the limit only rotation back toward the touchdown orientation is permitted, so the limit bounds departure without preventing recovery}.

The law \eqref{eq:imglaw} acts directly on the cue and needs no conversion to angle. \cdel{Two model quantities are nevertheless required by the terms below:}\cadd{Controller P below uses two model quantities:} the cue sensitivity $S=\partial I/\partial\theta$, which converts elapsed rotation into an equivalent change in the cue, and the rotation-to-force gain $g_\theta=\partial F_n/\partial\theta$ at fixed face position, which converts a candidate rotation into a force excursion. $S$ is estimated online by recursive least squares with forgetting ($\lambda=0.99$), initialized at \cdel{its calibrated value ($|S|=0.088$~rad$^{-1}$, Section~\ref{sec:char})}\cadd{a prior $|S|=0.05$~rad$^{-1}$ set before the calibration of Section~\ref{sec:char}}, with its sign fixed from calibration and updates \cdel{gated on appreciable rotation}\cadd{made only when the tilt has changed by more than $0.5^\circ$ over the preceding nine frames ($\approx0.3$~s)}.\cadd{ An underestimated $|S|$ under-compensates the delay, placing \fdel{P between full reconstruction and B2}\fadd{the reconstruction between none (as in B2) and full,} and biasing the reconstruction comparison toward no difference\fdel{, and in}\fadd{. In} the prediction\fadd{,} it slows the cue's decay and so overpredicts the excursion, \fdel{the same direction as}\fadd{as does} the gain bound.} $g_\theta$ is not estimated online: replaying the estimator on earlier phantom logs showed it is not identifiable from the controller's own rotations. It is instead held at a conservative bound $\bar g=40$~N/rad, above every value measured in Section~\ref{sec:char}.

\subsection{Compared Controllers}
All controllers share the force loop, \fdel{the rotation and excursion limits}\fadd{tilt limit}, and \fdel{the }cue.

\textbf{B1} applies no orientation correction, $\omega\equiv0$.

\textbf{B2} applies \eqref{eq:imglaw} to the most recent filtered cue $I_f$, with no delay reconstruction and no force constraint.

\textbf{P}\padd{, the proposed controller,} adds two terms. The first is \emph{delay reconstruction}. The cue was measured at capture time $t^c$ and the probe has rotated since, so the controller acts on
\begin{equation}
\hat I(t)=I_f(t)+S\bigl(\theta(t)-\theta(t^c-\tau_f)\bigr),
\label{eq:recon}
\end{equation}
with $\theta$ read from a $1$~kHz history buffer; the shift by $\tau_f$ compensates the filter's group delay. The second is a \emph{force-excursion constraint with fast-loop credit}\cadd{, which scales the commanded rate by $s\in[0,1]$}. At each new frame\cadd{,} the controller simulates\cadd{ for a candidate $s$}, over a horizon $T_h=100$~ms\fdel{ that covers the interval until the next frame}\fadd{, about one frame interval plus the capture-to-use delay (Section~\ref{sec:sensing})},
\begin{equation}
\dot\varepsilon=-a\,\varepsilon\cdelm{+}\cadd{-}\bar g'\omega,\quad
\dot{\hat I}=S\,\omega,\quad
\omega=s\,\operatorname{sat}\bigl(-K_I\sigma\hat I\bigr),
\label{eq:predsim}
\end{equation}
from $\varepsilon(0)=0$, where $\bar g'=\bar g/(1+KK_d)$ and $\varepsilon$ is the \cdel{part of $F_n-F_d$}\cadd{component of $e_F$} caused by the rotation\fadd{ over the horizon}. \cdel{Because the loop is linear, the current force error evolves the same way with or without rotation and cancels, so the constraint acts only on the excursion the rotation adds; removing the existing error is the force loop's task.}\cadd{Because \eqref{eq:arate} is linear, $e_F$ is the sum of \fadd{$\varepsilon$ and }the existing error, which decays at rate $a$ whether or not the probe rotates\fdel{, and $\varepsilon$}; the constraint therefore bounds only $\varepsilon$, and removing the existing error is left to the force loop.} The second equation lets the commanded rate fall as the cue is nulled within the horizon. With $u(s)=\max_{[0,T_h]}|\varepsilon|/\Delta F$\cadd{ the \fadd{normalized }simulated peak for a candidate $s$}, the \cdel{scale on the commanded rate}\cadd{applied scale} is
\begin{equation}
s=\arg\min_{s\in[0,1]}\;(1-s)^2-\mu\log\!\bigl(1-u(s)^2\bigr),
\label{eq:barrier}
\end{equation}
held until the next frame, with $\mu=0.05$ and $u(s)<1$. The scalar minimization is solved by a bounded golden-section search over the feasible range of $s\in[0,1]$. The first term is the cost of withholding correction: zero at full correction and largest at none. The second is a logarithmic barrier on the excursion budget $\Delta F$: negligible while the predicted excursion is small, so $s\to1$ when the correction is harmless, and unbounded as the prediction approaches $\Delta F$, so no minimizer can exceed it. The scale therefore varies continuously with the prediction rather than switching at the boundary\cdel{, which matters}\cadd{; continuity matters} because the prediction inherits the frame-to-frame variability of the cue.

The term $-a\varepsilon$ is what distinguishes P from a one-step constraint. A one-step constraint assumes the excursion accumulates unopposed, $\bar g\,\omega T_h$ over the horizon, whereas P credits the fast loop, which limits it to $(\bar g'\omega/a)(1-e^{-aT_h})$, so the correction is throttled less. The ablations change one setting each. \textbf{P\textsubscript{nb}} \cadd{(no barrier) }computes $s$ but does not apply it, isolating the constraint from reconstruction. \textbf{P\textsubscript{fix}} holds $S$ at its \cdel{calibrated value}\cadd{prior}. \textbf{P\textsubscript{1s}} is P\textsubscript{fix} with $a=0$ (and $\bar g'=\bar g$), the one-step (no-rejection) constraint. \textbf{B2\textsubscript{cap}} is B2 with $\omega_{\max}$ fixed at $0.20$~rad/s, near \fdel{the ceiling }$\omega^{*}$ \fdel{of }\eqref{eq:wstar}, testing whether a fixed cap reproduces the constraint.

\subsection{Activity of the Constraint}
At full rate, the constraint modifies the command \fadd{appreciably }only if the predicted excursion \fadd{approaches or }exceeds the budget,
\begin{equation}
\frac{\bar g'\,\omega_{\max}}{a}\bigl(1-e^{-aT_h}\bigr)\;\gtrsim\;\Delta F .
\label{eq:bindcond}
\end{equation}
The rotation limit $\omega_{\max}$ is the worst case, and at this operating point also the actual one, since the rate saturates immediately after \cdel{touchdown}\cadd{release}. Condition \eqref{eq:bindcond} can be checked before experiments: \fadd{any of }a fast force loop, a slow rotation limit, or a wide budget \fdel{each make}\fadd{makes} the constraint inactive, in which case P reduces to B2 with reconstruction, while compliant contacts, reduced force gains, and lower-bandwidth force loops enlarge the regime in which it acts. We use $\omega_{\max}=0.6$~rad/s and $\Delta F=0.5$~N, for which \eqref{eq:bindcond} holds with $\bar g$ (predicted excursion $1.64$~N at full rate and the median per-contact $K$), and report the realized scale $s$ \fdel{so that the comparison is not made at an operating point where the constraint never acts}\fadd{to confirm that the constraint acts}. The realized scale reached its minimum at the first evaluation after release, $s_{\min}=\result{0.27}$ (median over the \result{40} paired runs of P; range \result{0.265}--\result{0.276}), and \fdel{returned to $s=1$}\fadd{recovered to $s\ge0.99$} only after \result{2.6}~s, once the cue had largely converged, so the constraint acts over most of cue convergence rather than only at engagement. $\Delta F$ is a budget on the \fdel{rotation-induced excursion}\fadd{excursion a rotation adds within one horizon}, not a force-tracking band.

While the command is saturated, the constraint therefore acts as a ceiling on the rotation rate. Equating the predicted excursion to the budget gives the largest admissible rate,
\begin{equation}
\omega^{*}=\frac{a\,\Delta F}{\bar g'\bigl(1-e^{-aT_h}\bigr)},
\label{eq:wstar}
\end{equation}
which is $0.183$~rad/s for P and P\textsubscript{fix} at the median per-contact $K=1133$~N/m (Section~\ref{sec:char}), and $\Delta F/(\bar g T_h)=0.125$~rad/s for P\textsubscript{1s}, independent of $K$.

\noindent\emph{Proposition 1:} Under \eqref{eq:arate}, if $|g_\theta|\le\bar g$, $\varepsilon(0)=0$, and $|\omega(t)|\le\omega^{*}$ on $[0,T_h]$, then the rotation-induced error satisfies $|\varepsilon(t)|\le\Delta F$ on $[0,T_h]$. \emph{Proof:} By \eqref{eq:arate} with $|g'|\le\bar g'$, $|\varepsilon(t)|\le(\bar g'\omega^{*}/a)(1-e^{-at})\le\Delta F$ by \eqref{eq:wstar}.~$\square$

\fadd{Because each prediction starts from $\varepsilon(0)=0$, sustained rotation at $\omega^{*}$ lets the rotation-induced error approach $\Delta F/(1-e^{-aT_h})\approx1.0$~N under the gain bound.} Minimizing \eqref{eq:barrier} with $\mu=0.05$ at saturated demand places the \fdel{realized }rate below $\omega^{*}$, at $0.163$ and $0.108$~rad/s; the \fdel{measured}\fadd{logged} medians over the first $0.5$~s were \result{0.162} (P), \result{0.163} (P\textsubscript{fix}), and \result{0.108}~rad/s (P\textsubscript{1s}). \fdel{The ceiling of B2\textsubscript{cap} ($0.20$~rad/s)}\fadd{B2\textsubscript{cap}'s $0.20$~rad/s cap} lies $9\%$ above P's $\omega^{*}$ and \cdel{$20\%$}\cadd{$23\%$} above \fdel{its}\fadd{P's} realized rate.
\section{Characterization}
\label{sec:char}

All phantom experiments used a cylindrical gelatin phantom of radius $35$~mm, prepared from water, gelatin, and cornstarch (a speckle agent) in an $80\!:\!10\!:\!1$ weight ratio, with coupling gel at the interface. Scanner gain, dynamic range, time-gain compensation, speckle reduction, and adaptive gain were fixed for all runs, since the cue is comparable only within a fixed display configuration.\cdel{ Characterization was performed from the same taught start pose as the controller comparison of Section~\ref{sec:results}.}

\subsection{Contact Stiffness}
The three loading ramps of the tilt sweeps below gave $K=1371$~N/m (\fadd{median; }range $1279$--$1395$, $R^2$ $0.95$--$0.97$), which set the prior. Over the loading ramps of the comparison runs, the per-contact fit\fadd{, which the controller used,} gave $K=\result{1133}$~N/m (median; range \result{1068}--\result{1199}), the value \fdel{the controller used and }at which the rates of Section~\ref{sec:control} are evaluated; no run in the reported comparisons fell back to the prior.

\subsection{Tilt Sweep}
With the orientation loop disabled, the probe was stepped through face-pivot angles out to $\pm12^\circ$ and back while holding $15$~N, dwelling $3$~s at each angle, in three sweeps ($93$ dwells). The force loop returns $F_n$ to $F_d$ at every dwell (mean error below $0.01$~N), so $g_\theta$ is not observable as a force change; it appears as the depth change needed to maintain $15$~N. With $x$ the face depth along the touchdown beam axis, $y=F_n-Kx$ \cdel{is the force the tilt would have added at fixed depth}\cadd{removes the force explained by depth, leaving the tilt-dependent component up to a constant}, and $g_\theta=\mathrm{d}y/\mathrm{d}\theta$. No parametric angular-contact model is assumed: $g_\theta$ is estimated as the local slope of $y$ within fixed $4^\circ$ \fdel{angle }bands, after removing a slow drift\fdel{ in $y$ over the sweep} (gel relaxation, $-0.62$~N \fdel{over a}\fadd{per} $127$~s sweep) that the ascending and descending passes separate.

\emph{Rotation-to-force gain.}\cdel{ Because the surface is symmetric about its normal, $y$ is even about the tilt $\theta_n$ at which the face is normal to the surface, so $g_\theta$ vanishes there and changes sign across it.} Within the swept range $g_\theta$ \cdel{did not change sign}\cadd{kept one sign}: the banded slopes were all positive, between $11.5$ and $26.3$~N/rad (largest 95\% bound $33.8$~N/rad), a factor of $1.5$ below $\bar g$ ($1.2$ for the largest bound). The bound is thus verified only within $\pm12^\circ$; at larger tilts it is assumed.\cdel{ A quadratic fit places the vertex beyond the swept range ($\theta_n=-60^\circ$, 95\% CI $[-175,-39]^\circ$), so $\theta_n$ is reported only as lying beyond $-12^\circ$.}

\emph{Cue sensitivity.} Over the same dwells, the cue varied with tilt at $S=-0.088$~rad$^{-1}$ (95\% CI $[-0.096,-0.080]$, $R^2=0.85$), \cdel{which sets the calibrated value used by P\textsubscript{fix} and the initial value for online estimation. Extrapolated, its zero lies at $-30^\circ$, also beyond the swept range and on the same side as $\theta_n$, which is consistent with the start pose lying off the surface normal and the cue reporting it with the correct sign; the sweep does not resolve whether the two zeros coincide. The banded sensitivity did not steepen away from upright (mean $|S|$ for $|\theta|$ of $8$--$12^\circ$ was $0.9$ times that within $4^\circ$).}\cadd{$1.8$ times the prior used by the controllers. Under P the online estimate moved toward it from $-0.050$ during roughly the first half of each run and then held near a median of \result{$-0.072$}~rad$^{-1}$ (range \result{$-0.083$} to \result{$-0.064$} across runs), short of the calibrated value.}

\cdel{The control signal's age at use, from frame capture to the controller, had a median of $33.8$~ms, the confidence map running one frame behind the newest capture. The additional acquisition-to-display delay internal to the scanner was not resolved by a lag scan against known angle: the best-fit lag ($271$~ms, 95\% CI from $38$~ms to the $300$~ms edge of the scan) improved the fit by less than $0.001$ in $R^2$ over zero lag, and no delay beyond the measured age is compensated.}\cadd{\pdel{The control signal reached the controller a median of $33.8$~ms after the capture of its frame, the confidence map running one frame behind the newest capture, and was then held for up to one frame interval.}}
\section{Controller Experimental Validation}
\label{sec:results}

\subsection{Protocol}
The controllers of Section~\ref{sec:control} were compared on the phantom of Section~\ref{sec:char}, traversed along its axis for $100$~mm at $5$~mm/s with $F_d=15$~N and \cdel{image-based lateral centering active}\cadd{no lateral correction}. \cdel{The surface normal is constant along this path; the orientation demand arises at engagement, because every run starts from the same taught pose, which lies off the surface normal (Section~\ref{sec:char}), so}\cadd{The orientation demand arises at engagement: every run starts from the same pose, from which the cue reports a misalignment, so} the orientation loop rotates the probe as soon as it is released. By the time the cue reached $20\%$ of its initial value, the probe had rotated a median of \result{21.6}$^\circ$ under P (\result{20.0}--\result{26.9}$^\circ$ across the correcting conditions), toward the orientation of highest confidence, which need not coincide with the geometric normal. In every condition\cadd{,} the orientation loop is held off for $1$~s after touchdown, so that the force loop settles first\cadd{ and the touchdown transient is not confounded with rotation onset}, and all outcomes are measured from that release. All parameters\fdel{, including}\fadd{ (}$\omega_{\max}=0.6$~rad/s, $\Delta F=0.5$~N, $\bar g=40$~N/rad, the \cdel{calibrated}\cadd{prior} $S$, \fdel{the }force gains, \fdel{the }cue\cadd{ parameters}, and \fdel{the }scanner configuration\fdel{,}\fadd{)} were fixed before collection.

The seven comparisons of Table~\ref{tab:ablation} were collected as seven blocks of $20$ runs in interleaved ABBA order\cdel{, run as one automated sequence;}\cadd{;} each run is one experimental unit. Each run is paired with its time-adjacent partner of the other condition in the same \cdel{block (at most $300$~s apart)}\cadd{comparison}, giving $10$ pairs per comparison, balanced in order.\cdel{ The phantom shifted by about $1$~cm between some blocks, so comparisons are made only within blocks.}\cdel{ Three runs in which the scanner display froze, detected automatically as $15$ consecutive identical frames, were excluded without reference to condition and repeated at the end of the sequence; no other runs were excluded.}

The outcomes are the RMS force error over the first second after release, $E_1$, which captures the engagement transient, together with the peak error over the same second; the RMS force error from release to the end of the traverse, $E$; and the \pdel{alignment}\padd{cue-convergence} time $t_{I50}$, the time from release until the magnitude of the filtered cue first falls to half its initial value (its mean over the first $30$~ms). The residual cue, the median of $|I_f|$ over the last $5$~s, shows whether low cue asymmetry is maintained. Table~\ref{tab:ablation} reports paired differences. Confidence intervals quantify effect size and precision; no correction for multiple comparisons was applied.

\begin{table}[t]
\caption{Paired differences between conditions (A $-$ B), mean and \fadd{paired-$t$ }95\% CI, $n=10$ pairs each.}
\vspace{-3 mm}
\label{tab:ablation}
\centering
\footnotesize
\setlength{\tabcolsep}{3pt}
\begin{tabular}{lccc}
\toprule
A $-$ B & $\Delta E_1$ (N) & $\Delta E$ (N) & $\Delta t_{I50}$ (s) \\
\midrule
B1 $-$ B2 & $-1.95^*$ & $-2.07^*$ & $+6.66^*$ \\[-1pt]
 & {\scriptsize$[-2.08,-1.82]$} & {\scriptsize$[-2.19,-1.95]$} & {\scriptsize$[+5.91,+7.40]$} \\
B2 $-$ P & $+0.40^*$ & $+0.40^*$ & $-0.94^*$ \\[-1pt]
 & {\scriptsize$[+0.13,+0.67]$} & {\scriptsize$[+0.30,+0.51]$} & {\scriptsize$[-1.04,-0.84]$} \\
P\textsubscript{nb} $-$ P & $+0.49^*$ & $+0.54^*$ & $-0.93^*$ \\[-1pt]
 & {\scriptsize$[+0.29,+0.69]$} & {\scriptsize$[+0.33,+0.75]$} & {\scriptsize$[-1.00,-0.86]$} \\
B2 $-$ P\textsubscript{nb} & $+0.11\phantom{^*}$ & $+0.07\phantom{^*}$ & $+0.01\phantom{^*}$ \\[-1pt]
 & {\scriptsize$[-0.10,+0.32]$} & {\scriptsize$[-0.11,+0.25]$} & {\scriptsize$[-0.03,+0.05]$} \\
B2\textsubscript{cap} $-$ P & $+0.03\phantom{^*}$ & $+0.09\phantom{^*}$ & $-0.32^*$ \\[-1pt]
 & {\scriptsize$[-0.20,+0.25]$} & {\scriptsize$[-0.08,+0.26]$} & {\scriptsize$[-0.46,-0.19]$} \\
P\textsubscript{1s} $-$ P\textsubscript{fix} & $-0.96^*$ & $+0.03\phantom{^*}$ & $+0.59^*$ \\[-1pt]
 & {\scriptsize$[-1.20,-0.72]$} & {\scriptsize$[-0.16,+0.22]$} & {\scriptsize$[+0.43,+0.76]$} \\
P\textsubscript{fix} $-$ P & $-0.17^*$ & $+0.05\phantom{^*}$ & $-0.01\phantom{^*}$ \\[-1pt]
 & {\scriptsize$[-0.33,-0.01]$} & {\scriptsize$[-0.05,+0.16]$} & {\scriptsize$[-0.04,+0.02]$} \\
\bottomrule
\end{tabular}\\[2pt]
\raggedright\scriptsize $^*$95\% CI excludes zero.
\vspace{-5 mm}
\end{table}

\subsection{Results}
\begin{figure}[t]
\centering
\includegraphics{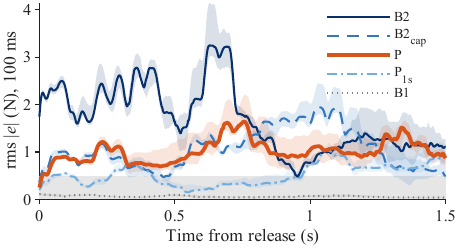}\\[-1pt]
\vspace{-2 mm}
{\footnotesize (a)}\\[2pt]
\includegraphics{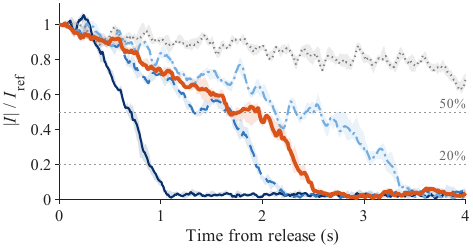}\\[-1pt]
\vspace{-2 mm}
{\footnotesize (b)}
\vspace{-2 mm}
\caption{Engagement and cue convergence after release ($t=0$)\cdel{, phantom, session~2.}\cadd{\fdelm{, phantom.}}\fadd{ on the phantom.}
(a)~Running RMS of force error (100\,ms centered window); shading \fdelm{marks $0.5$\,N}\fadd{marks the per-horizon budget $\Delta F=0.5$\,N}. (b)~\fdelm{Normalized image cue $|I|/I_\mathrm{ref}$}\fadd{Filtered cue magnitude over its mean in the first $30$\,ms, $|I|/I_\mathrm{ref}$}; dotted lines mark
\fdelm{the 50\% and 20\% thresholds defining $t_{I50}$ and $t_{I20}$}\fadd{50\% ($t_{I50}$) and 20\%}. Line styles as in (a). Curves are medians
over \fdelm{10 paired runs per condition with interquartile bands, each taken from its
own block}\fadd{the 10 paired runs of each condition in one block (B2: B2/P; P and B2$_\mathrm{cap}$: B2$_\mathrm{cap}$/P; P$_\mathrm{1s}$: P$_\mathrm{1s}$/P$_\mathrm{fix}$; B1: B1/B2), with interquartile bands}\fdelm{; they summarize runs, and the statistical comparisons are the paired differences in Table~\ref{tab:ablation}}\fadd{. Offsets between curves from different blocks are not paired comparisons; the paired comparisons are in Table~\ref{tab:ablation}}. P$_\mathrm{nb}$ and P$_\mathrm{fix}$ \fdelm{are omitted; they}\fadd{(omitted)} overlap B2 and P, respectively.}
\vspace{-5 mm}
\label{fig:overlay}
\end{figure}

\emph{Cost of correction.} B1 never rotates. Against B2 it lowers $E_1$ by $1.95$~N, the peak error in the first second by $5.98$~N, and $E$ by $2.07$~N, but it \cdel{leaves the probe misaligned}\cadd{does not actively reduce the cue}\fadd{ (Fig.~\ref{fig:overlay}b)}: its cue \cdel{reaches half its initial value only after a median of $7.4$~s, through drift along the traverse rather than correction, and its residual cue is}\cadd{\fdel{fell to half its initial value only by chance, after a median of $7.4$~s, as the probe traversed the phantom, and its residual stayed at}\fadd{reached half its initial value only after a median of $7.4$~s, without any correction, as the orientation of highest confidence shifted along the traverse, and ended at}} $0.009$ against $0.002$ for every correcting controller\fdelm{ (Fig.~\ref{fig:overlay}b)}. Orientation correction is not free: rotating the probe perturbs contact force through $g_\theta$. The remaining rows measure how that cost depends on \fdel{how fast the correction is applied}\fadd{the correction rate}.

\emph{Rate constraint.} Under B2 the probe rotates at the full rate from release\cdel{ (Fig.~\ref{fig:overlay})}. Relative to B2, P lowers $E_1$ by $0.40$~N, the peak error by $2.31$~N, and $E$ by $0.40$~N\cadd{\fdel{ (Fig.~\ref{fig:overlay}a)}}, and \pdel{aligns}\padd{its cue converges} $0.94$~s later (median $t_{I50}$ $1.62$ against $0.69$~s\cadd{; Fig.~\ref{fig:overlay}b}). Removing the constraint while keeping reconstruction and online estimation (P\textsubscript{nb}) raises $E_1$ by $0.49$~N, the peak by $2.74$~N, and $E$ by $0.54$~N, and shortens $t_{I50}$ by $0.93$~s. P\textsubscript{nb} and B2 do not differ resolvably on any outcome, \fdel{so the constraint accounts}\fadd{consistent with the constraint accounting} for the differences in both $E_1$ and $E$\fdel{, and reconstruction alone has}\fadd{ and with reconstruction alone having} no resolvable effect.\cadd{ B2 and P\textsubscript{nb} approached or reached the $31.5^\circ$ \fdel{excursion}\fadd{tilt} limit (median peaks \result{$29.7^\circ$} and \result{$31.6^\circ$}), whereas the rate-limited controllers stayed below \result{$28^\circ$}.}

\emph{Constraint versus a fixed cap.} B2\textsubscript{cap}, whose \fdel{ceiling}\fadd{cap} lies within $9\%$ of P's $\omega^{*}$ \eqref{eq:wstar}, does not differ resolvably from P in $E_1$ ($+0.03$~N) or $E$ ($+0.09$~N), and \pdel{aligns}\padd{its cue converges} $0.32$~s sooner (median $t_{I50}$ $1.32$ against $1.66$~s). Both limits act on the same saturated demand. P's realized rate (\fdel{$0.16$}\fadd{$0.162$}~rad/s, Section~\ref{sec:control}) lies \fdel{about $20\%$}\fadd{$19\%$} below the fixed \fdel{ceiling}\fadd{cap}, comparable to the $20\%$ difference in \pdel{alignment}\padd{cue-convergence} time, and the lower rate bought no resolvable reduction in force error; at this operating point the fixed cap is the better choice.

\emph{Fast-loop credit and adaptation.} Removing the credit (P\textsubscript{1s} against P\textsubscript{fix}) lowers the \fdel{effective rate}\fadd{ceiling} from $0.183$ to $0.125$~rad/s (realized\fadd{ rate} $0.163$ to $0.108$~rad/s). It lowers $E_1$ by $0.96$~N and the peak error by $2.70$~N, \fdel{leaves $E$ unchanged}\fadd{does not change $E$ resolvably}, and lengthens $t_{I50}$ by $0.59$~s (median $2.12$ against $1.61$~s). Holding $S$ at its \cdel{calibrated value}\cadd{prior} (P\textsubscript{fix}) lowers $E_1$ \fadd{marginally }by $0.17$~N and changes no other outcome resolvably\fdel{ ($t_{I50}$ within $0.04$~s)}.

\emph{Ordering.} Across blocks, median $t_{I50}$ follows the \cdel{rotation-rate ceiling}\cadd{realized rotation rate}: $0.65$--$0.69$~s for B2 and P\textsubscript{nb} ($0.6$~rad/s), $1.32$~s for B2\textsubscript{cap} ($0.20$), \fdel{$1.5$--$1.7$}\fadd{$1.52$--$1.66$}~s for P and P\textsubscript{fix} (\cdel{$\omega^*=0.18$}\cadd{$0.16$, the rate the barrier predicts below $\omega^*=0.18$}), and $2.12$~s for P\textsubscript{1s} (\cdel{$0.125$}\cadd{$0.11$}). Within every block that pairs clearly different \fdel{ceilings}\fadd{rate limits}, the lower \fdel{ceiling}\fadd{limit} had the lower $E_1$; this ordering across blocks is not itself a paired comparison.

\subsection{Volunteer Feasibility}
A single-subject popliteal scan tests whether the \cdel{frozen controllers}\cadd{same controllers, unchanged,} operate on living tissue, which adds spatially varying stiffness and curvature and subject motion (arterial pulsation, muscle activity, and limb drift) not represented by the phantom. It is reported as feasibility evidence and is not pooled with the phantom statistics.

P and B2 were each run six times in interleaved order and paired as on the phantom. Relative to P, B2 raised $E_1$ by $1.44$~N (95\% CI $[0.80,2.09]$), the peak error in the first second by $5.91$~N ($[3.32,8.50]$), and $E$ by $1.05$~N ($[0.82,1.29]$). Median $E$ was \result{2.02}~N under P and \result{3.09}~N under B2. On tissue the cue was dominated by noise: its magnitude at release (median $0.013$--$0.020$, against $0.08$ on the phantom) was comparable to its residual over the traverse ($0.017$--$0.019$), so convergence times on tissue do not measure alignment and are not reported. Before the reported runs, the probe placement was adjusted over several setup attempts, some of which ended at the force-limit safety stop, often when the rotation halted abruptly at the \fdel{excursion}\fadd{tilt} limit; these attempts are not analyzed\cdel{, and force estimates at this pose carry a further caveat}\fdel{, and the force estimate at this pose was not checked against a reference}. The \fadd{human }subject\fdel{, one of the authors,} gave informed consent under a protocol approved by the University of Alabama Institutional Review Board (protocol \fadd{ID 24-01-7230}). These data do not establish clinical accuracy, anatomical generalization, or diagnostic image quality.
\section{Discussion}
\label{sec:discussion}

The contribution is not the combination of image and force feedback, which is established~\rdelm{\cite{chatelain2015,jiang2020normal,akbari2021,zakeri2025,piedra2026}}\radd{\cite{jiang2020normal,akbari2021,zakeri2025}}, but a principled way to set how fast an image-guided correction may rotate the probe under force control. The coupling model gives the ceiling $\omega^{*}$ of \eqref{eq:wstar} in closed form, and the experiments \fdel{confirm it at three levels}\fadd{support it in two ways}.\fdel{ The realized rates matched the prediction within $0.001$~rad/s, with and without the fast-loop credit.} \pdel{Alignment}\padd{Cue-convergence} time followed the \fdel{ceiling}\fadd{realized rate}, and in every block pairing clearly different \fdel{ceilings}\fadd{rate limits}, the lower one had the lower engagement error. A fixed cap \fdel{at the predicted ceiling ($0.20$~rad/s, within $9\%$ of $\omega^{*}$) reproduced the force benefit of}\fadd{of $0.20$~rad/s, within $9\%$ of the predicted ceiling $\omega^{*}$, matched} the online predictive constraint\fadd{ in force error}, neither paired difference resolvable\fadd{ (upper 95\% bounds $0.25$ and $0.26$~N)}, while \pdel{aligning}\padd{converging} $0.32$~s sooner. The ceiling therefore carries the force benefit of the prediction and can be imposed directly: compute $\omega^{*}$ from the contact stiffness, the force-loop gains, the gain bound, \fdel{and }the budget\fadd{, and the horizon}, and cap the rotation rate at it. Delay reconstruction had no resolvable effect\cdel{ because it}\cadd{, although $S$ ran at \mbox{$57$--$82\%$} of its measured value; it also} had little to correct: under saturated demand the command ignores the cue's value, and in the proportional regime the \fdel{cue}\fadd{orientation} loop crosses over at $|S|K_I\approx3.5$~rad/s, where the \fadd{\pdel{measured }}$55$--$90$~ms capture-to-use delay\fadd{ (Section~\ref{sec:sensing}, with $\tau_f$)} costs only $10$--$20^\circ$ of phase\fadd{, not counting the scanner's \padd{unresolved }internal delay\pdel{, which was not resolved} (Section~\ref{sec:system})}. An online constraint can improve on the fixed cap where the demand falls below the ceiling or where stiffness or gain changes within a scan, which a single phantom did not test.

The rotation-to-force gain depends on surface geometry and contact state. On the cylindrical phantom, it kept one sign and stayed at or below $26.3$~N/rad over the $\pm12^\circ$ sweep (Section~\ref{sec:char})\fadd{, a narrower range than the rotations of Section~\ref{sec:results}}, and it was not identifiable online from the controller's own rotations.

The prediction \fdel{is therefore built from upper bounds rather than a model}\fadd{therefore uses a bound on the gain rather than a model of it}. The gain bound $\bar g=40$~N/rad exceeds every band gain measured on the phantom (largest 95\% bound $33.8$~N/rad), so the predicted excursion, and with it the admissible rate $\omega^{*}$, errs toward throttling. Where the true gain is smaller, the bound costs cue-convergence speed\fdel{ but cannot understate the force a rotation adds}. Because comparison-run stiffness varied over only $1068$--$1199$~N/m, the predicted dependence of $\omega^{*}$ on $K$ was validated only locally around this operating region.

Several limitations qualify these results. The contact-force estimate has no independent reference during operation and inherits the accuracy bound of Section~\ref{sec:system}, so all force results are stated in estimate space rather than as calibrated force, and they \cdel{include the arm dynamics the quasi-static estimator omits}\cadd{\fdel{contain the arm-dynamics torque that the quasi-static estimator does not model}\fadd{include the effect of the arm dynamics that the quasi-static estimator omits}} (Section~\ref{sec:system}). The budget $\Delta F=0.5$~N\pdel{ is comparable to that accuracy ($0.41$~N RMS), so it is enforced at a resolution the estimate cannot verify}\padd{, comparable to that accuracy ($0.41$~N RMS), constrains changes in the estimate rather than in the true force, whose excursion is not independently verified}. The confidence map's principal weakness is that a uniformly anechoic region presents no gradients to the random walk and may be assigned high confidence, so the cue reports asymmetry reliably but is not a contact detector. The volunteer scan exposed a second, related weakness. Over the curved popliteal fossa, one edge of the aperture lost acoustic coupling for sustained periods, and the loop did not null the resulting misalignment. The cue assumes that confidence differs between the two sides of the image mainly because of coupling; in heterogeneous tissue, attenuation and structure also lower and spread confidence on the coupled side, which likely diluted the asymmetry the decoupled edge should have produced. The \cdel{map is therefore insufficient as an alignment cue}\cadd{cue is therefore insufficient for alignment} on \fdel{nonhomogeneous}\fadd{heterogeneous} tissue. Surfaces on which the true gain approaches $\bar g$, such as stiffer tissue or sharper curvature, would need a different bound and were not tested.

On tissue the cue was noise-dominated, its magnitude at release comparable to its residual, yet P still lowered $E$ by $1.05$~N relative to B2, which makes the cue rather than the architecture the binding limitation on living tissue. The cue is low-pass filtered and the filter's group delay is compensated in \eqref{eq:recon}, but residual cue noise on tissue still converts directly into commanded rotation, and the \fdel{excursion}\fadd{tilt} limit bounds the consequence without addressing the cause. A confidence or cue measure that adapts to heterogeneous tissue rather than one tuned on homogeneous gelatin, or an explicit detector for edge decoupling, is the natural next step and is not attempted here. Finally, characterization is confined to homogeneous gelatin, and the volunteer scan is a single-subject feasibility test that establishes neither clinical safety nor anatomical generalization.
\section{Conclusion}
\label{sec:conclusion}

\pdel{We analyzed the coupling between a high-rate force loop using joint-torque-based estimation and a slower, delayed image-guided orientation loop in robotic ultrasound, and built a dual-rate controller that predicts the force excursion of a candidate correction from a conservative gain bound, credits the fast force loop with its rejection over the \cdel{frame interval}\cadd{prediction horizon}, and reconstructs the cue-convergence error from the frame's capture time. While the correction demand is saturated, the resulting constraint is a ceiling on the rotation rate that can be computed before each image-guided scan from identified \pdel{/contact-characterized quantities}\padd{system parameters}.}\padd{We analyzed the coupling between a high-rate force loop using joint-torque-based estimation and a slower, delayed image-guided orientation loop in robotic ultrasound. Under saturated demand, the proposed constraint, which credits the fast loop's rejection over the prediction horizon, reduces to a closed-form rotation-rate ceiling computable before each scan from identified system parameters.}

\pdel{On a curved phantom, \fdel{the realized rates matched that prediction, and }across seven paired comparisons the ceiling set a \cdel{speed--accuracy trade-off}\cadd{trade-off between \pdel{alignment}\padd{cue-convergence} speed and force error}: orientation correction carries a measurable force cost relative to holding orientation fixed, and lowering the rate reduced engagement force error at the cost of \pdel{alignment}\padd{cue-convergence} time. A fixed cap \fdel{at}\fadd{near} the predicted rate matched the controller on force and \pdel{aligned}\padd{converged} \cdel{sooner}\cadd{faster}, and delay reconstruction had no resolvable effect. On a single-subject popliteal scan, the rate-limited controller lowered the \fdel{force RMSE}\fadd{RMS force error $E$} by $1.05$~N relative to the unconstrained baseline, although the image cue was noise-dominated on tissue. Claims are restricted to the tested phantom and force range.}\padd{Across seven paired comparisons on a curved phantom, orientation correction carried a measurable force cost, and the ceiling set the trade-off between cue-convergence time and force error. A fixed cap near the predicted rate matched the online constraint in force error and converged faster, and delay reconstruction had no resolvable effect. For approximately stationary contact properties, the closed-form cap is therefore preferable to the online predictor because it provides comparable force regulation with lower implementation complexity. On a single-subject popliteal scan, the rate-limited controller lowered the RMS force error $E$ by $1.05$~N relative to the unconstrained baseline, although the image cue was noise-dominated on tissue. Future work includes developing an image cue robust to heterogeneous tissue and testing on more subjects and anatomies.}

\section*{Acknowledgment}
The authors acknowledge financial support from The University of Alabama faculty startup fund and the Alabama Transportation Institute. This work was also partly supported by the U.S. Department of Transportation University Transportation Centers Program through CR2C2, Grant No.~69A3552348304. Claude~\cite{anthropic_claude} was used under the authors' direction to assist with experiment planning, data analysis, code development, figure preparation, and manuscript drafting and editing, and ChatGPT~\cite{openai_chatgpt} to assist with manuscript review and editing. The authors built the experimental system and conducted and supervised all experiments; AI-assisted code was validated by testing on that system and by checking its outputs, and the authors take full responsibility for the content of this paper.
\balance
\bibliographystyle{IEEEtran}
\bibliography{references}
\end{document}